\documentclass[11pt,letterpaper]{llncs}
\usepackage{amsmath}
\usepackage{amssymb}
\usepackage{graphicx}
\usepackage{marvosym}
\usepackage{url}
\usepackage{fancyhdr}
\usepackage{xcolor}
\usepackage{authblk}
\usepackage{geometry}
\newcommand{\keywords}[1]{\par\addvspace\baselineskip
\noindent\keywordname\enspace\ignorespaces#1}

\fancypagestyle{firstpage}{\fancyhf{}
\setcounter{page}{1}
\fancyhead[C]{\small{}}
\fancyfoot[L]{}
\rfoot{\thepage}
}

\begin{document}

\title{\LARGE{Combining Discrete Wavelet and Cosine Transforms for Efficient Sentence Embedding}}
\author{\large{ Rana Salama\inst{1,}\inst{3}\and Abdou Youssef \inst{1} \and Mona Diab \inst{2}}}

\institute{\large \inst{} School of Engineering and Applied Science, George Washington University\and %
                      \inst{} Language Technologies Institute, Carnegie Mellon University \and %
                      \inst{} Faculty of Computers and Artificial Intelligence, Cairo University
                      }

\maketitle

\thispagestyle{firstpage}

\begin{abstract}
Wavelets have emerged as a cutting edge technology in a number of fields. Concrete results of their application in Image and Signal processing suggest that wavelets can be effectively applied to Natural Language Processing (NLP) tasks that capture a variety of linguistic properties. In this paper, we leverage the power of applying Discrete Wavelet Transforms (DWT) to word and sentence embeddings. We first evaluate, intrinsically and extrinsically, how wavelets can effectively be used to consolidate important information in a word vector while reducing its dimensionality. We further combine DWT with Discrete Cosine Transform (DCT) to propose a non-parameterized model that compresses a sentence with a dense amount of information in a fixed size vector based on locally varying word features. We show the efficacy of the proposed paradigm on downstream applications models yielding comparable and even superior (in some tasks) results to original embeddings.

\end{abstract}


\section{Introduction}

\label{sec:1}

Word embeddings are the basic building blocks for all recent NLP systems. They represent words in distributed dense real-valued vectors which geometrically encode the semantic and syntactic information of words in addition to their linguistic regularities~\cite{65}.
It has been noted that word embeddings have interesting algebraic characteristics capturing analogies and word relationships suggesting that dimensions along which words are aligned are correlated \cite{52.mikolov-etal-2013-linguistic}. These embeddings have a large mean vector and most of their salient features are located in a subspace of much fewer dimensions~\cite{67}. This representation has led many to view embeddings as a signal on which spectral techniques are applicable \cite{30.kayal-tsatsaronis-2019-eigensent}. Such spectral techniques transform data into a new space that captures their different characteristics and sketch new potentials for their usage and the information they convey. Recent spectral analysis in NLP include Discrete Cosine Transform (DCT)~\cite{1.almarwani-etal-2019-efficient} and Higher-order Dynamic Mode Decomposition (HODMD) EigenSent~\cite{30.kayal-tsatsaronis-2019-eigensent} embedding models. In this setting, a sentence is represented as a signal with transitional properties captured in the frequency domain using uncorrelated coefficients to encode a sentence. Such models proved to capture structural variation without losing on efficiency (comparable to averaging)~\cite{29.zhu2020sentence} and outperform more complex sentence embedding models~\cite{26.mikolov-etal-2018-advances}. However, these models tend to analyze embedding features along similar word embedding dimensions, on a vertical level (inter-word embedding, aka across all words), accumulating a limited number of base frequency coefficients and \underline{dropping} the rest, in addition to ignoring their position in the original domain, i.e., ignoring  intra-word spectral frequencies within individual word embeddings.
\\
\\Accordingly, in this paper we propose applying a successful method from image and signal processing, namely, Discrete Wavelet Transform (DWT), that proved to be very effective, fast, and space-efficient in Image Processing \cite{12.inbook}, Signal Processing \cite{8.STEPHANE2009481,4.91217} and Speech Recognition \cite{10.Trivedi2011SpeechRB}. 
\\
\\We primarily shed light on how DWT can be beneficial for NLP applications. We further adapt DWT to analyze data in terms of their spectral frequency with respect to their position in embedding representations to understand how data vary across different word embedding dimensions. Additionally, we recognize features that are highly varying and those that are nearly identical and can be combined into fewer dimensions. 
\\We posit DWT as an efficient method for word and sentence encoding. We explore DWT to selectively compress relevant information in a word embedding and effectively pack sentence embedding models with semantically condensed word representations. By applying DWT, we avert some of the observed drawbacks of previous spectral models that generate sentence embedding with exponentially increased dimensionality. 
\\
\\Our key contributions are: 
\\1) We propose a novel approach for leveraging DWT to word embeddings that selectively and efficiently compresses relevant salient information at different levels of detail;
\\2) We show the efficacy of our proposed adapted DWT approach on textual similarity as well as various downstream tasks using non-parameterized word embeddings;
\\3) We propose the novel approach of conjointly modeling DWT with DCT to encode variable length sentences into a fixed-size vector without hurting
performance. 

\begin{figure}
\centering
 \includegraphics[scale=0.6]{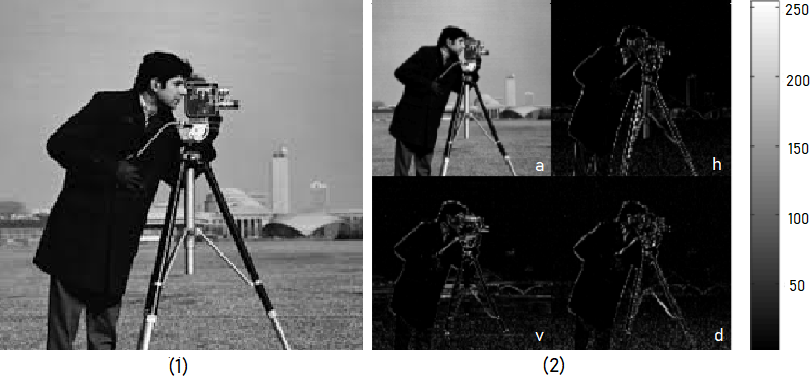}
  \caption{ \label{camerman} An example of Level-1 Wavelet transform using DWT to the cameraman image (1) and the corresponding transformed image (2) 
  The transform, (2), produces 4 sub-bands. The top left sub-band is a low resolution version of the original sub-sampled in both horizontal and vertical directions, while the horizontal, h, vertical, v, and diagonal, d, details represent the down-sampled residual versions of the original image.
}
\end{figure}

\subsection{Motivation}
\label{sec:2}

Wavelets Transforms (WTs) are mathematical transformation functions that switch between different levels of detail, or different resolution, to capture various insights about the data. As described in \cite{50}, WTs are like a “window", investigating data, with a large "window" would notice coarse features, while looking at the data with a small “window” would notice small features. Accordingly, WT analysis captures both the forest and the trees. In Image Processing, as shown in Figure \ref{camerman}, WT transforms the input iconic cameraman picture to four sub images: the top left sub image (or sub-band) captures a low resolution form of the main image (Approximation), comprising low-frequency content, and three other sub-bands comprising high frequency content (Details), that represent different edges present in the image along different directions, horizontal, vertical and diagonal. Accordingly, WTs have been effectively used for image compression~\cite{59,58}, and high-frequency components have been used for edge detection ~\cite{11.5089125,60.ZHANG20091265}. 
In signal processing, WTs have been used for signal compression \cite{63,62.doi:10.1080/14639230310001636499} \cite{8.STEPHANE2009481,4.91217}, speech recognition \cite{10.Trivedi2011SpeechRB} and noise filtering~\cite{61.8404418}. 
\\
\\{\bf From Image and Signal Processing to NLP}. WTs can provide considerable
insight to NLP as it had in Image and Signal Processing. WTs are expected to provide more insight about the data which should be of great value to NLP. WT can be used to analyze words and sentence representations based on spectral frequency of their variation across time, i.e. where they appear in the vector. Basically, WT can be used to build models that generate arbitrarily good approximations of word representations eliminating irrelevant details, hence generating compressed representations. Also, WT can be used to derive high-frequency representations, which capture details, nuances and contrasts between different features, resulting in equally minimized representations suitable for applications where contrast and subtle nuances are of special value.

\subsection{DCT Sentence Embedding}
\label{sec:3}

Sentence Embedding is considered important for transferring knowledge to downstream tasks in NLP \cite{41.422b29624234f9441cab6dd9c918e86ab7}. Recent advances in sentence embedding proposed using DCT coefficients to compress word vectors considering order and structure of words in a sentence \cite{1.almarwani-etal-2019-efficient}. 
Later studies proved the efficiency of DCT embedding to capture semantic regularities \cite{29.zhu2020sentence} and demonstrated how it outperforms more complex sentence embedding models  \cite{26.mikolov-etal-2018-advances}. \\However, DCT sentence embedding is mainly based on the idea of concatenating the first \textit{K} DCT coefficients, resulting in a sentence vector of size \textit{Kd}, where \textit{d} is the word vector dimension, which constitutes a very long representation as the number of coefficients increases. For example, the proposed model in \cite{1.almarwani-etal-2019-efficient} achieved its best performance over a number of tasks using K=6, which results in a sentence vector of size 1800 when using a word vector size of 300. This increased sentence embedding dimension and its impact on the performance of NLP tasks has been a recent research question \cite{dim} suggesting that the increased size of a sentence embedding is usually sub-optimal.

\begin{figure*}
\centering
 \includegraphics[scale=0.9]{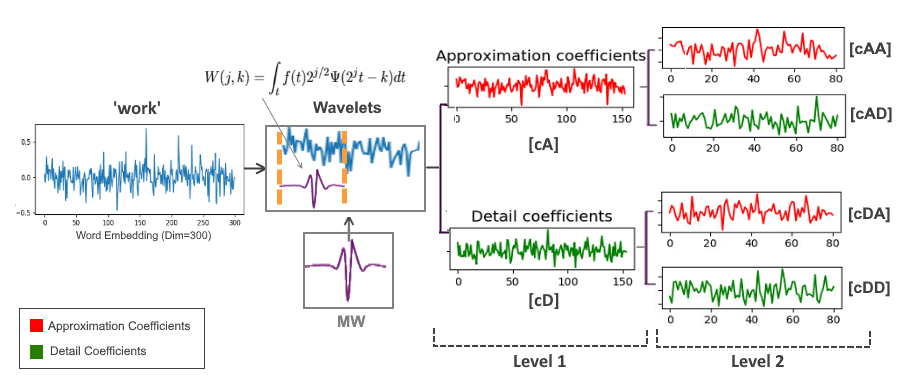}
  \caption{ \label{wav} Plotting word embedding of the word ‘work’ and its coefficients across 2 levels of WT. cD and cA represent Detail and Approximation coefficients derived from Level-1 of the transform. cDA and cDD indicate Level-2 Approximation and Detail coefficients derived from applying DWT on the Level-1 Detail coefficients. cAA and cAD indicate Level-2 Approximation and Detail coefficients derived from applying DWT on the Level-1 Approximation coefficients.}
\end{figure*}

\section{Wavelet Transforms}

WT presents a time-frequency analysis tool that can effectively transform a given function \textit{f(t)} into another domain making certain features more amenable to study and analyze\cite{20.1191457}. In our context, \textit{f(t)} represents the word embedding vector. A WT is applied by translating and shifting a convolution-like function called the Mother Wavelet (MW) $\Psi$(t) over \textit{f(t)}. MWs are like a microscope zooming in to see details or zooming out to ignore details and see an approximation of the data.\\
\\Basically, WT calculates the correlation between the MW and the word vector at different segments of position. Higher correlation indicates more similarity. Using these correlation values, WT efficiently generates pairs of low-pass (low-frequency/high-correlation) and  high-pass (high-frequency/low-correlation) filters. The low-pass filters capture Approximation coefficients(cA), and the high-pass filters capture Detail coefficients(cD). The output of a filter pair is usually down-sampled by two. This filtering+downsampling process can be applied on multiple levels recursively as shown in Figure \ref{wav}. First level transformation will produce cA and cD coefficients downsampled by 2. The second level transformation will further transform cA into new approximation and details coefficients, denoted cAA and cAD, that are further downsampled by 2 in size. Similarly, cD will be transformed into cDA and cDD. 
\\
\\It should be noted that although WT can be used as continuous and discrete transforms, in the context of this paper we only consider Discrete Wavelets Transforms, DWT, which is more suitable for NLP data, which is discrete, that is, word embeddings are finite-dimensional vectors (of real numbers) rather than continuous (i.e., analog) signals.
\\
\\Formally, given a MW $\Psi$(t) that can be scaled by factor j and shifted by k, we get the conjugate of $\Psi$(t) according to the following equation:

\[\Psi _{j,k}(t)= 2^{j/2} \Psi (2^j t-k)\]  
The general DWT, denoted by W(j,k), is then given by

\[ W(j,k) = \int_{t} f(t)2^{j/2} \Psi (2^jt-k)dt\]
To convert data back from the wavelet domain to the original domain, an inverse DWT can be applied. 
\\
\\Note that those definitions are continuous convolutions. In the discrete case as in NLP, discrete convolution counterparts are used. 
Therefore, WT can be compared to Convolutional Neural Networks (CNNs)~\cite{74} in that they both use sliding-window filters and downsampling (pooling). The difference is that in CNNs, the filters are learned from the training data, whereas in WT the filters are designed, not learned. Though less tuned to the data, WT are more computationally efficient in terms of time and space. In fact, some WT algorithms, such as Mallat’s pyramid, are linear in time and space~\cite{5.article}.
\\
\\There are many families of WT functions available in the literature and used as MWs including: Haar, Symmlets, Coiflets, Daubechies, and Biorthogonal, to name a few~\cite{20.1191457}. 
Yet as a proof of concept in this paper, and in the interest of space efficiency, we will only be using a subset of the MWs in our experiments. \footnote{For a more detailed explanation of WT theory,  refer to~\cite{73}~\cite{20.1191457}~\cite{68.brunton_kutz_2019}.}
\\
\\Comparing DWT to other transforms such as Discrete Cosine Transforms (DCT), DWT allows good localization in both time and frequency domains permitting spectrum analysis of the data, in addition to its spectral behavior in time. DWT allows analyzing related features in word embeddings by their frequency with respect to where they occur. In DCT, feature frequencies are analyzed based on their variation regardless of their location in a word embedding. As a result, DWT can give more analytical insights about the data being transformed.
DWT also proved to yield higher compression ratios at comparable fidelity, as well as inherent scaling \cite{4.91217}.

\begin{figure}[htb]

\centering
 \includegraphics[scale=0.9]{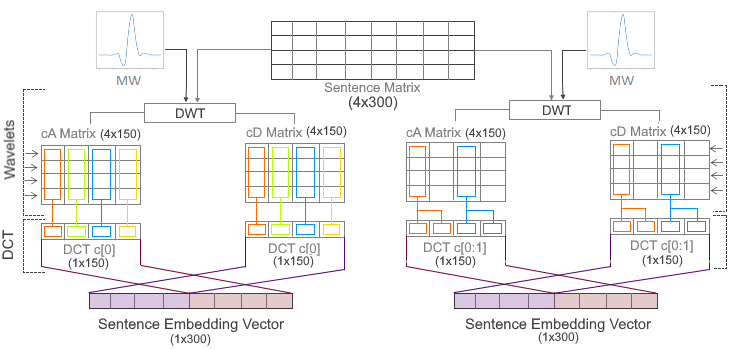}
  \caption{ \label{model}Proposed Wavelet-DCT sentence embedding model applied to a sentence of 4 words represented as a matrix. Level-1 DWT and DCT-1 are used as demonstrated on the left side, while the right side demonstrates using Level-1 DWT with DCT-2 by skipping every-other coefficient in DWT coefficients matrix to generate sentence embedding representation.
 }
\end{figure}

\section{Wavelet-DCT Sentence Embedding} \label{sec:3}

We propose a novel method of applying wavelets conjointly with DCT for sentence embedding. WT can be used to reduce the dimensionality of a word vector retaining important features with respect to local variation, while DCT coefficients are further considered to 
preserve word order in a sentence. 
\\
\\Given a sentence of \textit{N} d-dimensional word vectors, $w_1,..,w_N$, we initially stack the words in an 
$N\times d$  
matrix. The sentence matrix is then DWT-transformed row-wise, where every row represents a word,  for L levels (L can be 1, 2, ..., n) to compress word embeddings. 
This transformation will break the sentence matrix into $2^{L}$ matrices each of size $N\times\frac{d}{2^{L}}$, representing the WT coefficients. For L=1, we will have two $N\times \frac{d}{2}$ matrices, one for cA and one for cD. For L=2, we will have four matrices of size $N\times \frac{d}{4}$, namely, cDD, cDA, cAD and cAA.
Since we want the sentence embedding to further inherit the details and coefficients of its words, we will keep all DWT-coefficients, but we will further summarize their pattern along different words using a DCT. Applying DCT to summarize DWT coefficients is like the crème de la crème for the N words in a sentence.
\\
\\Accordingly, the DWT-coefficients matrices will be transformed, column-wise, using DCT. We finally encode a sentence vector of size $d$ by concatenating either 1 or 2 DCT coefficients, which proved to be sufficient empirically and also consistent with observations in \cite{1.almarwani-etal-2019-efficient}, for final sentence representation. As mentioned in \cite{1.almarwani-etal-2019-efficient} we will be using $K$ to denote the number of coefficients and $c[0:K-1]$ for their actual values.
When the number of DCT coefficients is 1; $K=1$, the DCT-DWT-coefficients matrices will be of size 
$1\times\frac{d}{2^{L}}$ each, concatenating them results in a final sentence embedding of size d. 
\\
\\However, when $K=2$, we generate our final sentence embedding vector by using a simple \emph{trick} that skips every other DWT coefficient before applying DCT. This will further reduce a DWT-coefficient matrix into 
$N\times\frac{d}{2^{L+1}}$. As a result, selecting 2 DCT coefficients and concatenating them will still result in a sentence vector of size $d$. This trick proved, empirically, not to affect the information preserved by these coefficients but keep the sentence embedding dimension unchanged. 

\section{Evaluation of DWT Efficacy}

We initially investigate the efficacy of applying DWT to word embeddings. We explore DWT coefficients and selectively using them as new compressed embeddings. We evaluate the efficacy of the new embeddings to capture and compress semantics using semantic similarity tasks. We further evaluate their effectiveness in a number of downstream tasks to evaluate their efficacy extrinsically. 
\\
\\In this evaluation, we will consider the following DWT coefficients as embeddings; Level-1 Detail ($cD$) and Approximation ($cA$) coefficients. Additionally, we extend these coefficients with Level-2 coefficients (performed on Level-1 Approximation and Detail coefficients as shown in Figure \ref{wav}), $cD+cAD$ and $cA+cDA$, to enrich the word representation with more information from subsequent transformation levels.

\subsection{Experimental Setup}

\subsubsection{Embeddings} 
For our experiments we used a number of base embeddings to demonstrate the capabilities of DWT. In this context we use GloVe \cite{85} and FastText\cite{26.mikolov-etal-2018-advances} for word embeddings with various dimensions (100, 200, 300) for a comprehensive evaluation. 

\subsubsection{Sentence Embedding Evaluation} 
For sentence embeddings, we use the SentEval toolkit~\cite{28.DBLP:journals/corr/abs-1803-05449} for evaluation. For all downstream tasks, we employed multi-layer perceptron (MLP) classifiers based on the setup outlined in SentEval. 

\subsection{Intrinsic Evaluation}

We evaluate DWT embeddings and their encoded semantics in two evaluation tasks: Word Similarity and Concept Categorization.
For Word Similarity, we use the following datasets : SimLex-999 \cite{23.DBLP:journals/corr/HillRK14}, MEN\cite{25.10.5555} and WS353\cite{ws353}.

\subsubsection{Baseline}
We set the original embeddings, i.e. without transformation, as our baseline: F-BASELINE and G-BASELINE1, corresponding to FastText and GloVE embeddings.
In GloVE, we consider another baseline, G-BASELINE2, which is the original GloVE embedding with 50\% less dimensions to compare the reduced DWT embeddings to the original embeddings of the same size, i.e. for a GloVE embeddings of size 100, the transformed DWT embeddings will be of size 50, so we consider GloVE embedding of size 50 as another baseline.
\\
\\Our experiments show that Level-1 DWT embeddings, cA and cD, with 50\% reduction in size, surpass the performance of the baselines, G-BASELINE1 and G-BASELINE2, as depicted in Table~\ref{simG}.
Given the original dimension of 100, for the SimLex dataset, cA outperforms G-BASELINE1 by 8\%. Moreover, cA exhibits superior performance compared to G-BASELINE2 and increases performance by 11\% even though they share the same size. This empirically demonstrates that DWT embeddings capture the underlying semantics conveyed in the embeddings and effectively encoding them into fewer dimensions.
While for WS353 and MEN datasets, cD beats both baselines, illustrating that for different datasets, every set of coefficients captures different aspects about the data.
\\Furthermore, for dimension of 200, cA outperforms specifically for the SimLex and MEN datasets, emphasizing how various coefficients capture information within the embeddings as the dimension size increases.
\\
\\In our subsequent experiment, we utilize FastText embeddings with 300 dimensions as the baseline, F-BASELINE. Furthermore, we explore Level-2 DWT embeddings, specifically, we enrich the Level-1 cA coefficients with the approximation coefficients from the Level-2 cD transform, cDA. Similarly, we examine the Level-1 cD coefficients alongside the Level-2 detail coefficients for Level-1 approximations, cAD.
As shown in Table \ref{simF}, at a compression of 25\% dimensionality reduction, cD+cAD yields comparable results to F-BASELINE for all 3 datasets. For the SimLex dataset, the cD coefficients effectively encode the semantics, and the addition of coefficients from subsequent layers (cD+cAD) didn't improve the performance. \footnote{In this context, cDD will represent the details of the details of Level-1 and hence won't add much information} 
However, for the MEN and WS353 datasets, combining the coefficients (cD+cAD) achieves performance comparable to the F-BASELINE embeddings, with a slight reduction of at most 1\% in performance.

\begin{table}[htb]
\caption{Word Similarity Evaluation using GloVe Embeddings}
\label{simG} 
\centering
\begin{tabular}{|l|c|c|c|c|} 
\hline
 & \textbf{Dim} & \textbf{SimLex} & \textbf{WS353} & \textbf{MEN} \\ 
\hline
\textbf{G-BASELINE1} & \textbf{100} &0.12 & 0.46 & 0.57 \\ 
\hline
\textbf{G-BASELINE2} & \textbf{50} &0.09 &0.42 & 0.53 \\ 
\hline
cD & 50 & 0.11 & \textbf{\textcolor{red}{0.50}} & \textbf{\textcolor{red}{0.58}} \\ 
\hline
cA & 50 & \textbf{\textcolor{red}{0.2} } & 0.44 & 0.57 \\ 
\hline
\hline
\textbf{G-BASELINE1} & \textbf{200} & 0.13 & \textbf{0.48} & 0.59 \\ 
\hline
\textbf{G-BASELINE2} & \textbf{100} & 0.12 & 0.46 & 0.57 \\ 
\hline
cD & 100 & 0.13 & 0.40 & 0.48 \\ 
\hline
cA & 100 & \textbf{\textcolor{red}{0.17}} & \textcolor{red}{0.47} & \textbf{\textcolor{red}{0.62}} \\
\hline
\end{tabular}

\vspace{3mm}

\small{ Spearman Rank Order Correlation (SPC) Results on SimLex-999, WS353 and MEN datasets; using GloVe-Twitter27B embeddings with dimensions 100 and 200 compared to Level-1 DWT coefficients on word similarity tasks. In addition to G-BASELINE2 which represents GloVE embedding with 50\% less dimensions. cD and cA correspond to the embeddings yielded at Level-1 DWT transform. Best results are in bold and best results per experimental condition are in red.
}
\end{table}
%


\begin{table}[h]
\caption{Word Similarity Evaluation using FastText Embeddings}
\label{simF} 
\centering
\begin{tabular}{|l|c|c|c|c|} 
\hline
 & \textbf{Dim} & \textbf{SimLex} & \textbf{WS353} & \textbf{MEN} \\ 
\hline
\textbf{F-BASELINE} & \textbf{300} & \textbf{0.5} & \textbf{0.79} & \textbf{0.83} \\ 
\hline
cD & 150 & \textbf{\textcolor{red}{0.5}} & 0.73 & 0.79 \\ 
\hline
cA & 150 & 0.49 & 0.74 & 0.79 \\ 
\hline
cD+cAD & 225 & \textbf{\textcolor{red}{0.5}} & \textcolor{red}{0.78} & \textcolor{red}{0.82} \\ 
\hline
cA+cDA & 225 & 0.49 & 0.75 & \textcolor{red}{0.82} \\
\hline
\end{tabular}
\\
\end{table}
\noindent We further evaluate DWT embeddings using the Concept Categorization task that groups words in different categories based on semantic clusters~\cite{54.baroni-etal-2014-dont}. Such  models have been proven effective in downstream NLP tasks~\cite{22.wang_wang_chen_wang_kuo_2019}.
We evaluate DWT embeddings using the AP~\cite{55.article}, BM~\cite{57.bm.5555/2387636.2387658} and BLESS datasets~\cite{56.bless.inproceedings}. We use FastText as original embeddings. As illustrated in Table~\ref{cg}, cA and cD embeddings yield comparable and even superior results compared to F-BASELINE with a dimensionality reduction of 50\%.

\begin{table}[htb]
\caption{ Concept Categorization Evaluation}
\label{cg} 
\centering
\begin{tabular}{|l|c|c|c|c|} 
\hline
 & \textbf{Dim} & \textbf{AP} & \textbf{BM} & \textbf{BLESS} \\ 
\hline
BASELINE & 300 & \textbf{0.70} & 0.47 & 0.86 \\ 
\hline
cD & 150 & \textbf{\textcolor{red}{ 0.70}} & 0.46 & \textbf{\textcolor{red}{0.87}} \\ 
\hline
cA & 150 & \textbf{\textcolor{red}{0.70}} & \textbf{\textcolor{red}{ 0.49}} & 0.82 \\
\hline
\end{tabular}
\end{table}

\vspace{3mm}

\noindent Furthermore, we study DWT embeddings qualitatively, we take a snippet of different word pairs and their cosine similarity measures using FastText embeddings, cA and cD embeddings as shown in Table~\ref{wordpair}. The results show that, on average, cD performs as well as original word embeddings. However, for cA, some interesting results hold: For pairs like, 'dog-cow' and 'happy-cry', cA embeddings tend to outperform with a big margin both the original embeddings and the cD coefficients.
This could typically indicate that these two words have a high correlation between their Approximation coefficients and basically capture a different type of relation, for example, 'dog' and 'cow' are both animals, and 'happy' and 'cry' are both emotions. It could also indicate that the low-pass filtering (which produces the cA's) preserves and highlights the commonality between 'dog' and 'cow' (their animal nature), and between 'happy' and 'cry' (their emotional nature), leading to higher similarity, whereas the high-pass filtering (which produces the cD's) eliminates/reduces those kinds of commonalities, but preserves certain other connections that yield the same (or even slightly better) similarity value  as/than the original embeddings, but lower similarities than those produced by cA. 
\\
\\On the other hand, for the 'boy-girl' and 'woman-girl' pairs, cA decreases the similarities drastically, while cD does not, indicating that the subtle nuanced connection/contrast (of gender or age) got lost in the low-pass filtering but was preserved by the high-pass filtering. These curious observations call for further investigation into the working of the low-pass filtering and high-pass filtering of WT, a subject for future research.

\begin{table}
\centering
\caption{Similarity measures for different word pairs }\label{wordpair}
\begin{tabular}{|l|c|c|c|} 
\hline
\textbf{Word Pairs} & \textbf{Word Vector} & \textbf{cD} & \textbf{cA} \\ 
\hline
'boy'-'girl' & 0.77 & 0.78 & 0.57 \\ 
\hline
'dog'-'cow' & 0.39 & 0.39 & 0.89 \\ 
\hline
'woman'-'girl' & 0.65 & 0.67 & 0.49 \\ 
\hline
'happy'-'cry' & 0.28 & 0.37 & 0.91 \\
\hline
\end{tabular}
\end{table}

\noindent We additionally captured the 5-nearest neighbors for a set of randomly selected words, as shown in Table \ref{5nn}. cD and cA were able to capture some linguistic phenomena that the original word embeddings missed. For example, 'Cat' was not one of the 5-nearest neighbors for 'cat'. Synonyms like 'contented' are closer to happy as opposed to 'unhappy' in the original representation. When using cA, we capture more synonyms of 'happy' like 'thrilled' and 'overjoyed', and when using cD, we capture arguably better relational similarity  to 'Italy' such as 'sicily' and 'levorno', which are Italian cities. 

\begin{table*}
\caption{5-nearest cosine similar words}
\label{5nn}
\centering

\resizebox{\textwidth}{!}{
\begin{tabular}{|l|c|c|c|} 
\hline
\textbf{Word} & \textbf{Word Vector} & \textbf{cD} & \textbf{cA} \\ 
\hline
'cat' & \begin{tabular}[c]{@{}c@{}}'cats', 'kitty', 'kitten',\\~'feline', 'kitties'\end{tabular} & \begin{tabular}[c]{@{}c@{}}cats', 'kitten', 'kitty',\\~'feline', 'Cat'\end{tabular} & \begin{tabular}[c]{@{}c@{}}'cats', 'kitty', 'kitten',\\~'feline', 'kitties'\end{tabular} \\ 
\hline
'happy' & \begin{tabular}[c]{@{}c@{}}'happpy', 'unhappy', 'hapy',\\~'contented', 'happier'\end{tabular} & \begin{tabular}[c]{@{}c@{}}'happpy', 'contented', 'Happy',\\~'hapy', 'unhappy'\end{tabular} & \begin{tabular}[c]{@{}c@{}}'thrilled', 'happpy', 'unhappy',\\~'overjoyed', 'hapy'\end{tabular} \\ 
\hline
'Italy' & \begin{tabular}[c]{@{}c@{}}'rome', 'italy.', 'spain',\\~'france', 'europe'\end{tabular} & \begin{tabular}[c]{@{}c@{}}'spain', 'sicily', 'italy', \\'livorno', 'europe'\end{tabular} & \begin{tabular}[c]{@{}c@{}}'rome', 'italy', 'france',\\~'switzerland', 'germany'\end{tabular} \\
\hline
\end{tabular}}

\end{table*}

\subsection{Extrinsic Evaluation}

For extrinsic evaluation, DWT embeddings are applied in the same settings of intrinsic evaluation to the following downstream classifications tasks through SentEval toolkit~\cite{28.DBLP:journals/corr/abs-1803-05449}: sentiment classification on Movie Reviews (MR),  Stanford Sentiment Treebank (SST2, SST5)~\cite{32.pang-lee-2004-sentimental}, product review (CR)~\cite{33.Hu2004MiningAS}, subjectivity classification (SUBJ)~\cite{34.pang-lee-2004-sentimental},
opinion polarity classification (MPQA), question type classification (TREC)~\cite{36.10.1145/345508.345577}, paraphrase identification (MRPC)~\cite{37.dolan-etal-2004-unsupervised}, STS12 semantic similarity task~\cite{sts12},  and entailment classification on the SICK dataset (SICK-E)~\cite{12.inbook}.
We use DWT embeddings to encode sentences then feed the encoded sentences to the aforementioned tasks. 

\subsubsection{Baseline}
In this evaluation, we consider averaging as the baseline sentence encoding (AVG), using FastText word embeddings. We further consider other baselines to demonstrate the efficacy of DWT embeddings including: (1) FastText embeddings with random 50\% dimensionality reduction, denoted as random pooling, under the assumption of no correlation between dimensions. This involves the random elimination of 50\% of the features in a word embedding.(2) An embedding model that is based on a dimensionality reduction method, namely PCA(Principal component analysis), as developed in~\cite{67} and denoted as (PCA+PPA). This embedding  combines PCA dimensionality reduction embeddings with a post-processing algorithm (PPA)~\cite{82} to construct effective word embeddings of lower dimensions. 
\\
\begin{table*}
\caption{Extrinsic Results}
\label{extrinsic}
\resizebox{\textwidth}{!}{
\begin{tabular}{|l|c|c|c|c|c|c|c|c|c|c|c|} 
\hline
 & \multicolumn{1}{l|}{} & \multicolumn{5}{c|}{\textbf{Sentiment Analysis}} & \multicolumn{1}{c}{\textbf{Inference}} & \multicolumn{2}{c|}{\textbf{Paraphrase}} & \multicolumn{1}{c}{\textbf{SUBJ}} & \textbf{TREC} \\ 
\hline
\textbf{Embedding} & \textbf{Dim} & \textbf{MR} & \textbf{CR} & \textbf{SST2} & \textbf{SST5} & \textbf{MPQA} & \textbf{SICK-E} & \textbf{MRPC} & \multicolumn{1}{l|}{\textbf{STS12}} &  &  \\ 
\hline
AVG & $300$ & $78.3$ & $79.6$ & $\textbf{84.13}$ & $44.16$ & $87.94$ & $\textbf{79.5}$ & $74.43$ & $58.3$ & $92.33$ & $83.2$ \\ 
\hline
AVG(Random Pooling) & $150$ & $58.83$ & $63.58$ & $56.07$ & $28.82$ & $69.61$ & $56.59$ & $67.07$ & $8$ & $69.78$ & $33$ \\ 
\hline
AVG(PCA+PPA) & $150$ & $75.52$ & $78.2$ & - & $41.4$ & $86.18$ & $75.06$ & $73.39$ & $53.34$ & $90.96$ & $75.2$ \\ 
\hline
AVG(PCA+PPA) & $200$ & $77.18$ & $79.76$ & - & $43.48$ & $86.64$ & $76.76$ & $72.93$ & $53.76$ & $91.6$ & $77.4$ \\ 
\hline
AVG(cA) & 150 & $\textcolor{red}{\textbf{78.57} }$ & $80.85 $ & $80.12 $ & $43.6 $ & $85.61 $ & $78.41 $ & $73.22 $ & $58.01 $ & $90.01 $ & $82 $ \\ 
\hline
AVG(cD) & 150 & $76.61_coif15$ & $79.26 $ & $80.62 $ & $43.53 $ & $86.41 $ & $78.46 $ & $73.62 $ & $57.71 $ & $91.97 $ & $78.6 $ \\ 
\hline
AVG(cD+cAD) & 225 & $78 $ & $80.69 $ & $80.94 $ & $44.7 $ & $87.89 $ & $78.53 $ & $73.86 $ & $\textcolor{red}{\textbf{58.34} }$ & $92.12 $ & $81.8 $ \\ 
\hline
AVG(cA+cDA) & 225 & $77.4 $ & $\textcolor{red}{\textbf{81.22} }$ & $\textcolor{red}{82.43 }$ & $\textcolor{red}{\textbf{45.52} }$ & $87.36 $ & $78.49 $ & $73.49 $ & $57.94 $ & $91.96 $ & $83.2 $ \\ 
\hline
AVG(cD+cAD+cAAD) & 263 & $78.38 $ & $81.09 $ & $81.55 $ & $43.94 $ & $\textcolor{red}{\textbf{87.95} }$ & $\textcolor{red}{78.89 }$ & $\textcolor{red}{\textbf{74.5} }$ & $\textcolor{red}{\textbf{58.34} }$ & $\textcolor{red}{\textbf{92.34} }$ & $\textcolor{red}{\textbf{84.8} }$ \\
\hline
\end{tabular}
}

\vspace{3mm}

\small {Best Classification accuracy (except STS12, the metric is Pearson correlation \%) results on various classification tasks. The Baseline AVG is the simple original word embedding averaging. AVG(PCA+PPA) is the average of a PCA based embedding proposed in ~\ cite{82} with dimensions 150 and 200~\ cite{67}. AVG(cD) represents the average  of Level-1 Detail coefficients, AVG(cD+cAD) is the average of Level-1 Detail coefficients concatenated to Level-2 Detail coefficients as derived from Level-1 Approximation coefficients. Similarly for AVG(cD+cAD+cAAD) and AVG(cD+cAD+cAAD). We also illustrate the MW used per condition per task in italics. The best overall results are shown in bold. Best results per condition are shown in red.}
\end{table*}
\\In Table~\ref{extrinsic}, the overall results are comparable to the baselines AVG for most tasks with significant reduction in dimensionality, yet DWT outperforms in CR, SST5, MPQA, MRPC, STS12, SUB and TREC. In this experiment we further consider Level-3 DWT embeddings as shown in the table. For SUBJ, Level-3 yields the best results with a compression rate of 12.5\%. For the sentiment classification tasks we note performance comparable to AVG for MR and MPQA, while outperforming AVG for CR and SST5 at Level-2 cA+cDA with compression rate of 25\%. We also note that in the SST2 task, none of the conditions beat AVG. Similarly, for the entailment task of SICK-E none of the conditions surpassed AVG. 
However, for the Paraphrase tasks of MRPC and STS12 as well as the question classification task TREC, Level-3 representations yielded comparable performance to AVG with a compression rate of 12\%.  
\\
\\In general, Level-2 AVG(cA+cDA) and Level-3 AVG(cD+cAD+cAAD) yield the best results compared to other conditions except for the MR task where AVG(cA) yields the highest result (78.57\%) followed by the Level-3 condition at an accuracy of 78.38\%. 
Furthermore, if the embedding size is a major concern, observe that with cA or cD alone (at half the embedding size relative to the original), the performance is comparable to the AVG baseline in nearly all the tasks considered.
Hence, we conclude that DWT presents an effective balance between efficiency (compactness) and accuracy, and an effective data-size reduction method with hardly any adverse effect on accuracy.

\section{Evaluating DWT with DCT Sentence Embedding}
We evaluate the effectiveness of DWT by studying the application of DWT to the DCT sentence embedding model. We argue that DWT can encapsulate ample and pertinent information within word embeddings, leading to an improved representation for sentence embeddings without increasing dimension size nor compromising performance.

\subsubsection{Baseline}
In this evaluation, we consider a number of sentence embedding models as a baseline including: 
\begin{enumerate}
    \item 
Similar spatial models, namely, DCT~\cite{1.almarwani-etal-2019-efficient} with 1 and 2 coefficients to be consistent with our proposed model,  and EigenSent model~\cite{30.kayal-tsatsaronis-2019-eigensent}\footnote{Embeddings generated using https://github.com/DeepK/hoDMD-experiments}.
EigenSent sentence embeddings are composed by keeping the top \textit{m} dynamic modes resulting in a sentence embedding of size \textit{md} ignoring correlation and information encoded within a word embedding for every word in the sentence.
\item
Other top performing non-parametrized sentence embedding models: P-Means~\cite{40.DBLP:journals/corr/abs-1803-01400} and VLAWE~\cite{39.ionescu2019vector} to comprehend our result analysis.
\end{enumerate}
For a fair comparison, we use multi-layer perceptron (MLP) classifiers based on various settings including number of hidden states (in [0, 50, 100, 200]) and dropout rates (in [0, 0.1]) considering the same settings in 
\cite{1.almarwani-etal-2019-efficient} that yielded their best results for DCT-based sentence embedding.

\subsection{DWT-DCT Experimental Results}
The results on different tasks demonstrate the power of applying DWT to word embeddings. The combined DWT-DCT outperforms the baseline AVG in all tasks with a significant margin in some tasks. 
Comparing our approach to DCT alone, as shown in Table \ref{res}, our combined model yields better results. It can be shown that Level-1 or Level-2 DWT with 1 coefficient, DWT1-DCT[0] and DWT2-DCT[0], generally outperforms DCT with 1 coefficient, DCT c[0], in all task except MR which is comparable.
Similarly with 2 coefficients, DWT1-DCT[0:1], applying DWT to DCT performs better in most of the tasks, except for SUBJ which is comparable, with 50\% reduction in size.
It is worth mentioning that in CR, our DWT-DCT embeddings outperforms the best result achieved in~\cite{1.almarwani-etal-2019-efficient}, which is 80.08, when $K=5$, which results in a final DCT sentence embedding of size 1500.
In SICK-E, DCT embedding model achieves a comparable result when $K=4$, which results in a final DCT sentence embedding of size 1200. Additionally, in SST2, DCT embedding didn't beat the AVG baseline, while our model outperformed it. 
As for EigenSent, our model significantly surpasses the others, with the same embedding size, across all tasks.
Compared to other non-parametrized models, our model surpasses VLAWE in all tasks. As opposed to P-Means, our model outperforms or is comparable to the other models, except for SICK-E and STS16, yet our results are convenient given the fact that our embedding dimensionality is 12 times lower in magnitude.
\\
\\Our observations suggest that DWT seems to be complementary to DCT as we observe better performance. Additionally, Level-1 DWT is sufficient to pack a sentence embedding with relevant information.
\\
\\Overall, the combined model is robust and efficient, yielding results that are better than or comparable to the state of the art models across a variety of standard tasks. 

\begin{table*}
\caption{Experimental results from applying DWT to DCT sentence embeddings}
\label{res}
\centering
\resizebox{\textwidth}{!}{
\begin{tabular}{|l|c|c|c|c|l|c|c|c|c|c|c|c|c|} 
\hline
 &  & \multicolumn{5}{c|}{\textbf{Sentiment Analysis}} & \multicolumn{1}{c}{\textbf{Inference}} & \multicolumn{4}{c|}{\textbf{Paraphrase}} & \multicolumn{1}{l}{\textbf{SUBJ}} & \multicolumn{1}{l|}{\textbf{TREC}} \\ 
\hline
\textbf{Model} & \textbf{Dim} & \textbf{MR} & \textbf{CR} & \textbf{SST2} & \textbf{SST5} & \textbf{MPQA} & \textbf{SICK-E} & \textbf{MRPC} & \textbf{STS12} & \textbf{STS16} & \textbf{STSB} &  &  \\ 
\hline
\textbf{\# Samples} & - & 11k & 4k & 70k & 19.5k & 11k & 10k & 5.7k & 8.1k & 3.9k & 15k & 10k & 6k \\ 
\hline
\textbf{AVG } & 300 & 78.3 & 79.6 & 84.13 & 44.16 & 87.94 & 79.5 & 74.43 & 58.0 & 64.0 & 69.26 & 92.33 & 83.2 \\ 
\hline
\textbf{p-means} & 3600 & 78.3 & 80.8 & 84 & \multicolumn{1}{c|}{-} & \textbf{89.1} & 83.5 & 73.2 & 54 & \textbf{67} & 72 & 92.6 & \textbf{88.4} \\ 
\hline
\textbf{VLAWE} & 3000 & 77.7 & 79.2 & 80.8 & \multicolumn{1}{c|}{-} & 88.1 & 81.2 & 72.8 & - & - & - & 91.7 & 87 \\ 
\hline
\textbf{EigenSent} & 300 & 70.26 & 73.16 & 72.54 & 36.97 & 69.15 & 71.34 & 70.43 & 36 & - & - & 85.73 & 54.2 \\ 
\hline
\textbf{DCT c[0]} & 300 & 78.45 & 79.81 & 83.53 & \multicolumn{1}{c|}{44.57} & 88.36 & 78.91 & 72.93 & 58.3 & 64.1 & 71.5 & 92.79 & 84.8 \\ 
\cline{1-13}
\textbf{DWT1-DCT[0]} & 300 & 78.38 & \textit{81.14} & \textcolor{red}{\textit{\textbf{84.68}}} & \textit{46.73} & \textcolor{red}{88.36} & \textit{79.5} & \textit{73.33} & \textit{58.6 } & \textit{64.5} & \textit{71.6 } & \textcolor{red}{\textit{\textbf{92.8}}} & 84.6 \\ 
\hline
\textbf{DWT2-DCT[0]} & 300 & 78.31 & \textcolor{red}{\textit{\textbf{81.27}}} & \textit{83.86} & \textit{45.93} & 88.35 & \textit{79.58} & \textit{73.51} & \textcolor{red}{\textit{\textbf{58.73}}} & \textcolor{red}{\textit{64.8}} & \textcolor{red}{\textit{\textbf{72.3}}} & \textcolor{red}{\textit{\textbf{92.8}}} & \textit{85} \\ 
\hline
\textbf{DCT c[0:1]} & 600 & 78.15 & 79.84 & 83.47 & 46.06 & 87.76 & 79.64 & 72.81 & 50.4 & 57.4 & 71.2 & 92.61 & 88.2 \\ 
\hline
\textbf{DWT1-DCT[0:1]} & 300 & \textcolor{red}{\textit{\textbf{78.57}}} & \textit{80.66} & \textit{83.91} & \textcolor{red}{\textit{\textbf{46.74}}} & \textit{87.93} & \textcolor{red}{\textit{\textbf{80.9}}} & \textcolor{red}{\textit{\textbf{74.5}}} & \textit{50.8} & 57.4 & \textit{71.4} & 92.51 & \textcolor{red}{\textit{\textbf{88.4}}} \\
\hline
\end{tabular}

}
\small{ Results as opposed to DCT embedding, DCT c*, as reported in \cite{1.almarwani-etal-2019-efficient}. Our model names use the convention ‘DWT$l$-DCT[$k$]’ to indicate $l$-Level DWT followed by DCT where coefficients $k$ of the DCT output are taken for the final representation; e.g., DWT1-DCT[0] is used for Level-1 of DWT and c[0] DCT coefficient. 'Dim' reflects the sentence embedding vector size. In STS12, STS16 and STSB Pearson correlation coefficients(\%) is used. The best overall results are shown in bold. Best results per condition are shown in red. Results where the DWT-DCT sentence embedding based model exceeds the DCT model are shown in italic.
}
\end{table*}

\section{Conclusion and Discussion}

The results on different tasks demonstrate the power of our method. Our combined DWT and DCT outperforms the baseline AVG (in the majority of cases, by a significant margin) across all tasks. It outperforms  p-means and VLAWE as well, despite their several orders of magnitude larger dimensionality for both p-means and VLAWE. Comparing DCT to our combined models, the combined models yielded better results especially compared to DCT[0:1] and DCT[0:2] (except for TREC) despite having larger dimensionality for DCT.  We also note that DWT seems to be complementary to DCT as we observed better performance in the combined models except for MR where the best results are the same, and for TREC where DCT[0:2] outperforms all the combined models.
Overall, our combined models are robust and efficient, yielding results comparable or even outperforming the state of the art of non-parameterized results across a variety of standard tasks.

\subsection{Mother Wavelets}

While we simplified our paper by omitting the specific details of MW and its application in each task, it remains a crucial element of WT. In our experiments, we explored the use of Coiflets, Daubechies, and Symlets wavelets at different scales; i.e. how stretched is the MW across the data.
We observed that Coiflets, on average, demonstrate strong performance across various tasks and scales of expansion. This suggests that the Coiflets wavelets fits well the structure of the embedding data.
\\
\\Intuitively, we surmise from our experiments that frequent words will tend to transfer better with wide scaled wavelets to capture approximation details more efficiently. Conversely, less frequent, specialized words in a given context may yield better results when transformed using smaller-scaled wavelets. The selection of the best MW and scale to be used requires more research and we plan to investigate this more in our future work.

\subsection{Correlation}
Although we did not explicitly prove that the dimensions in the same embedding are correlated, the results of applying DWT suggests that there exists a correlation between these dimensions. If no correlation existed, DWT would never achieve comparable results to original embeddings' performance.

\subsection{Computational Complexity}
The computational complexity of applying DWT to embedding is typically linear. The general complexity is often expressed in terms of the dimension of the embedding, denoted as $N$ and depends on the number of levels in the DWT. If the transform has $L$ levels, the overall complexity is $$O(L\times N)$$ Accordingly, using DWT for embeddings analysis and compression is computationally efficient. 

\subsection{Conclusion}
In this paper we explored the effectiveness of applying DWT to word embeddings to \textit{selectively} reduce word embeddings into approximated or detailed coefficients representations by exploring frequency and space analysis of a word embedding, retaining the same performance at 50\% to 75\% of the word vector original dimension size.
The generated DWT word embedding postulates that different sets of coefficients capture different semantic aspects of a word embedding. Our intrinsic and extrinsic evaluations for the efficacy of applying DWT in the context of NLP suggest that DWT has significant potential for \textit{efficiently} modeling word and sentence embeddings. 
\\We further use the resulting word embeddings to generate a DWT-DCT based sentence-embedding. The proposed embedding method not only yields comparable or even better performance than original embedding models, it also has the added advantage of significantly reducing sentence vector size, whittling it down to salient info for the task. 
Finally, our model is able to outperform vector averaging for the SST Task which to date is the dominating model among non-parametric models. 

\bibliographystyle{plain}
\bibliography{bls_final}

\vspace{2cm}

\end{document}